\PassOptionsToPackage{prologue,dvipsnames}{xcolor}
\documentclass[sigconf]{acmart}
\usepackage{times}
\usepackage{comment}
\usepackage{soul}
\usepackage{url}
\usepackage[utf8]{inputenc}
\usepackage{graphicx}
\usepackage{amsmath}
\usepackage{amsthm}
\usepackage{booktabs}
\usepackage{algorithm}
\usepackage{dsfont}
\usepackage{makecell}
\usepackage[noend]{algpseudocode}
\usepackage{enumitem}
\usepackage[switch]{lineno}
\usepackage[T1]{fontenc}
\usepackage{tgcursor}
\usepackage{multirow}
\AtBeginDocument{%
  }

\setcopyright{acmlicensed}
\copyrightyear{2018}
\acmYear{2018}
\acmDOI{XXXXXXX.XXXXXXX}

\acmISBN{978-1-4503-XXXX-X/18/06}

\begin{document}

%%
%% The "title" command has an optional parameter,
%% allowing the author to define a "short title" to be used in page headers.
\title{Graph-DPEP: Decomposed Plug and Ensemble Play for Few-Shot Document Relation Extraction with Graph-of-Thoughts Reasoning}

%%
%% The "author" command and its associated commands are used to define
%% the authors and their affiliations.
%% Of note is the shared affiliation of the first two authors, and the
%% "authornote" and "authornotemark" commands
%% used to denote shared contribution to the research.
\author{Tao Zhang}
\affiliation{%
  \institution{University of Illinois at Chicago}
  \city{Chicago}
  \state{Illinois}
  \country{USA}
}
\email{tzhang90@uic.edu}

\author{Ning Yan}
\affiliation{%
  \institution{Futurewei Technologies Inc.}
  \city{San Jose}
  \country{USA}}
\email{nyan@futurewei.com}

\author{Masood Mortazavi}
\affiliation{%
  \institution{Futurewei Technologies Inc.}
  \city{San Jose}
  \country{USA}}
\email{masood.mortazavi@futurewei.com}

\author{Hoang H. Nguyen}
\affiliation{%
  \institution{University of Illinois at Chicago}
  \city{Chicago}
  \state{Illinois}
  \country{USA}
}
\email{hnguy7@uic.edu}

\author{Zhongfen Deng}
\affiliation{%
  \institution{University of Illinois at Chicago}
  \city{Chicago}
  \state{Illinois}
  \country{USA}
}
\email{zdeng21@uic.edu}

\author{Philip S. Yu}
\affiliation{%
  \institution{University of Illinois at Chicago}
  \city{Chicago}
  \state{Illinois}
  \country{USA}
}
\email{psyu@uic.edu}

%%
%% By default, the full list of authors will be used in the page
%% headers. Often, this list is too long, and will overlap
%% other information printed in the page headers. This command allows
%% the author to define a more concise list
%% of authors' names for this purpose.
\renewcommand{\shortauthors}{Trovato et al.}

%%
%% The abstract is a short summary of the work to be presented in the
%% article.
\begin{abstract}
Large language models (LLMs) pre-trained on massive corpora have demonstrated impressive few-shot learning capability on many NLP tasks. Recasting an NLP task into a text-to-text generation task is a common practice so that generative LLMs can be prompted to resolve it. However, performing document-level relation extraction (DocRE) tasks with generative LLM models is still challenging due to the structured output format of DocRE, which complicates the conversion to plain text. Limited information available in few-shot samples and prompt instructions induce further difficulties and challenges in relation extraction for mentioned entities in a document. In this paper, we represent the structured output as a graph-style triplet rather than natural language expressions and leverage generative LLMs for the DocRE task. Our approach, the Graph-DPEP framework is grounded in the reasoning behind triplet explanation thoughts presented in natural language. In this framework, we first introduce a ``decomposed-plug" method for performing the generation from LLMs over prompts with type-space decomposition to alleviate the burden of distinguishing all relation types. Second, we employ a verifier for calibrating the generation and identifying overlooked query entity pairs. Third, we develop "ensemble-play", reapplying generation on the entire type list by leveraging the reasoning thoughts embedded in a sub-graph associated with the missing query pair to address the missingness issue. Through extensive comparisons with existing prompt techniques and alternative Language Models (LLMs), our framework demonstrates superior performance on publicly available benchmarks in experiments.
\end{abstract}

%%
%% The code below is generated by the tool at http://dl.acm.org/ccs.cfm.
%% Please copy and paste the code instead of the example below.
%%



%%
%% Keywords. The author(s) should pick words that accurately describe
%% the work being presented. Separate the keywords with commas.
%% A "teaser" image appears between the author and affiliation
%% information and the body of the document, and typically spans the
%% page.

%\received{20 February 2007}
%\received[revised]{12 March 2009}
%\received[accepted]{5 June 2009}

%%
%% This command processes the author and affiliation and title
%% information and builds the first part of the formatted document.
\maketitle

\section{Introduction}
\begin{figure}[ht!]  
    \centering
    \includegraphics[width= 0.5\textwidth]{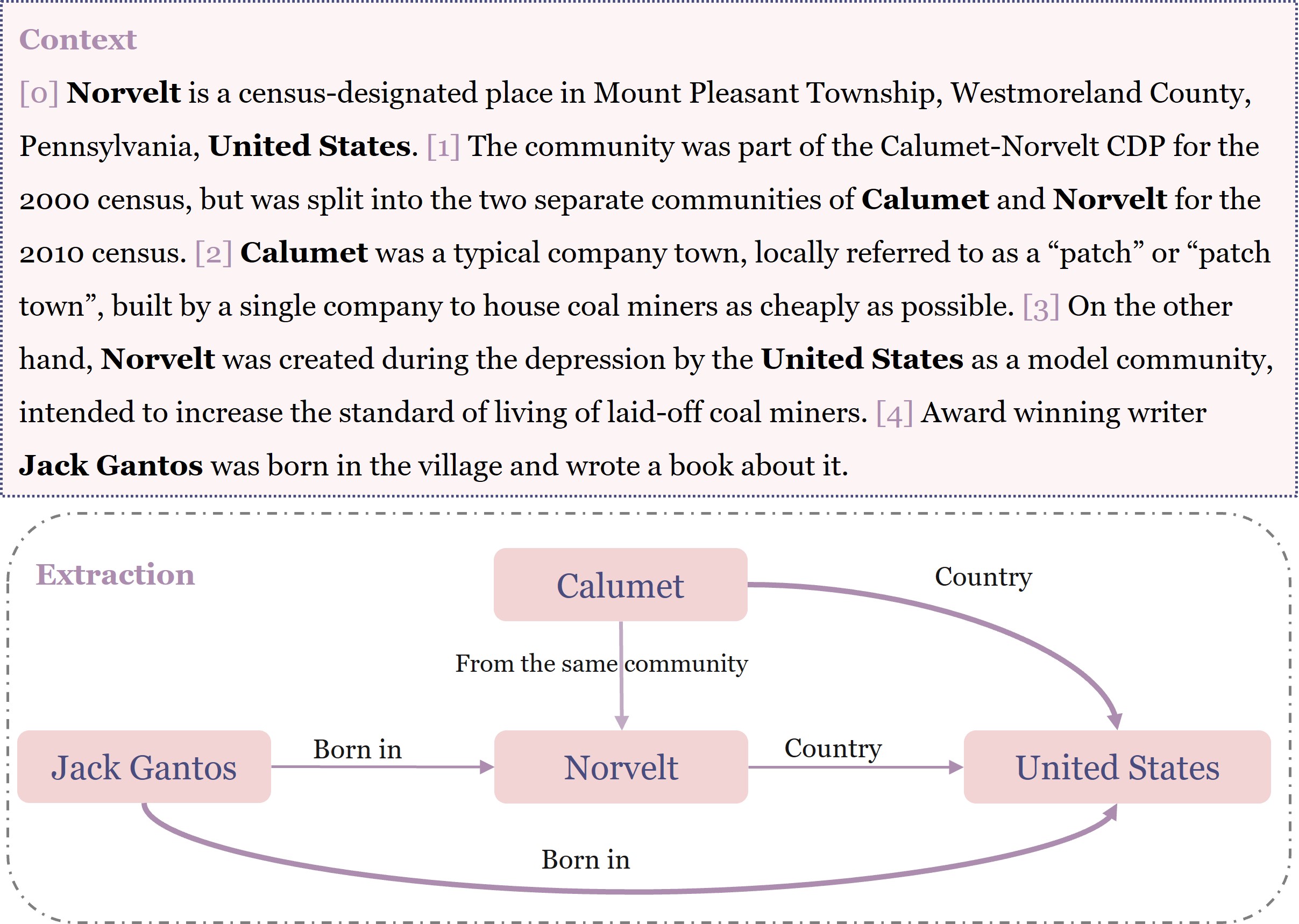}
    \vspace{-0.1in}
    \caption{An example of document relation extraction. }
    \label{fig: example}
\end{figure}
Document-level relation extraction (DocRE) extracts relations among multiple entity pairs in a document, representing a more realistic and challenging task than sentence-level extraction \cite{ji2017distant,alt2020probing}. In DocRE, an entity can have multiple mentions scattered throughout a document, and relationships between entities can appear in multiple different sentences. We illustrate a running example in Figure \ref{fig: example}. All the entities that occurred in the context are marked in bold. These relations can be identified by intra-sentence hints, like "Jack Ganto was born in Norvelt", which can be found in \textit{Sent\#4}. But "Calumet is in the country, the United States", which should be reasoned from \textit{Sent\#0} and \textit{Sent\#1}, because \textit{Sent\#0} indicates that Norvelt is in the United States and \textit{Sent\#1} indicates Calumet and Norvelt are from the same community. Most traditional transform-based DocRE models \cite{ma2023dreeam,tan2022document,yan2021unified} filter information in a lengthy document by retrieving relevant sentences as the supporting evidence to identify the relation for a given query entity mention pair. The idea of supporting evidence retrieval inspires us to make reasoning thoughts when we recast the traditional DocRE into a generation task.

Large language models (LLMs) like ChatGPT present remarkable success in generative and reasoning tasks. Reformulating the traditional NLP tasks into text-to-text generation formats to perform LLMs attracts intensive focus, especially under low-resource scenarios. For DocRe, annotation on evidence sentences is expensive, which restrains traditional model ability in real-world scenarios. Therefore, we leverage LLMs as the few-shot learner to address few-shot DocRE in a generative manner. 

However, existing LLMs are mostly pre-trained on unstructured data, which leads to poor performance when dealing with tasks that require a structured format. 
%The challenge is to adapt DocRE into an LLM generative task necessitating natural-language-information description to interact with LLMs.
Prior studies \cite{wadhwa2023revisiting,li2023codeie} utilize graph-like triplet sets, code-style frames, and similar approaches. As we also utilize triplet notation for describing extracted relations, we can establish a semantically rich graph structure to store information in a format readable by large language models (LLMs).
On the other hand, even with a reasonable number of few-shot examples, LLMs still face challenges in generating graph-structured data, and the outputs from LLMs remain error-prone.
Moreover, compared with regular relation extraction \cite{wadhwa2023revisiting,li2023codeie}, DocRE suffers from extraction on a large label space, which poses a significant challenge in response time and quality of LLMs. %natural language response engineering.
Even given several relation extraction examples for in-context learning, LLMs are susceptible to the risk of misinterpreting labels within the vast number of possibilities. 

In this paper, we investigate the end-to-end few-shot DocRE problem via a generative model and propose Graph-DPEP, a decomposed-plug and ensemble-play framework that allows self-verification on generation under relation graph-of-thoughts reasoning.
To assist LLMs in distinguishing intensive labels when transferring a classification task into a generation task, the decomposed-plug component processes an LLM to make a generation on each single type. What's more, we inspect the generation results with a verifier module for concerns such as repetition, irrelevance, incompleteness, and missing query pairs. The verifier incorporates calibration principles to refine the generation, addressing issues except for missingness, which is identified for subsequent adjustments to compensate for performance losses in this aspect. Reapplying LLM on the missing pairs can be facilitated by an association sub-graph that is relevant to entities in the missing query pairs. So, we employ the association sub-graph as the graph-of-thoughts to further aid reasoning on missing pairs' relations from the entire type list rather single one in the decomposed plug, so-called ensemble play.  

We summarize our contributions as follows:
\begin{itemize}
    \item To our knowledge, we are the first to transfer DocRE from a classification task to a generation task by constructing a decomposed method for prompt engineering to address the huge relation type space.
    
    \item We compare our approach with the best existing prompting techniques available to us and our Graph-DPEP method achieves the most promising recalls on the whole type space, especially for infrequent types.
    
    \item We employ the latest LLMs for evaluation of the effectiveness of the prompt design in Graph-DPEP. The experiment results show the advantages of our decomposed-plug and ensemble-play method with a self-correctable verifier.
\end{itemize}

\section{Related Work}
Transformer-based models have proven more effective than graph-based approaches \cite{zeng2020double,zeng2021sire,xu2021document} for Document-level Relation Extraction (DocRE) by leveraging long-distance token dependencies via Pre-trained Language Models (PLMs). Xiao \shortcite{xiao2021sais} introduces intermediate steps to enhance reasoning in relation extraction. Zhang \shortcite{zhang2021document} employs a U-Net structure to capture both local and global entity dependencies. Zhou \shortcite{zhou2021document}  uses localized contextual pooling to focus on relevant tokens for each entity pair. Tan \shortcite{tan2022document} implements axial attention and two-hop reasoning to better capture relational dependencies. Generative methods depict extraction results in a generative model vary among existing works, such as linearization on entities for named entity recognition subtasks\shortcite{yan2021unified}, code-style of output in both named entity recognition and relation extraction in CODEIE \cite{li2023codeie} and relation triplet form for standard sentence-level RE in \cite{wadhwa2023revisiting}.

%\noindent\textbf{Few Shot In-Context Learning}
%As a new paradigm, In-Context Learning does not require parameter updates and directly performs predictions on the pre-trained language models. The
%provided demonstration examples in the prompt follow the same format, which is usually written in natural language templates. By concatenating
%a query question with the demonstrations in the prompt, LLMs can learn from the given examples and predict the query question. Previous research \cite{liu2022makes,lu2022fantastically} has shown that the number of demonstrations and the order of the demonstrations can inﬂuence the In-Context Learning performance.

Small calibration models can improve LLM performance. Cobbe et al. \cite{cobbe2021training} use a verifier to assess the correctness of multiple candidate solutions in math word problems. Welleck et al. \cite{welleck2022generating} introduce a self-correction mechanism via iterative refinement. Han et al. \cite{han2023pive} propose the PiVe framework, which leverages iterative verification to enhance graph-based generation in LLMs.

\section{Graph-Enhanced Plug-and-Play DocRE}
\label{sec: method}
\begin{figure*}[ht!]
    \centering
    \includegraphics[width= \textwidth]{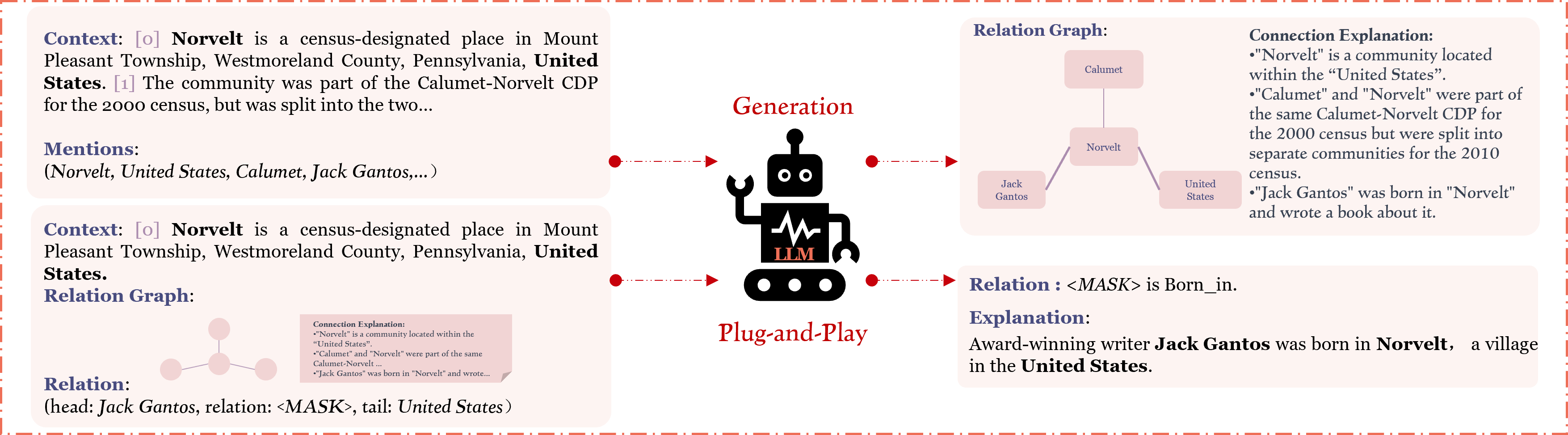}
    \vspace{-0.1in}
    \caption{Generative DocRE.}
    \label{fig: Framework}
\end{figure*}

In this section, we first formulate the document-level relation extraction (DocRE) task as a structure prediction problem. Subsequently, we
describe how we recast this structure prediction from a lengthy text
task into a generative task (Section \ref{subsec: Task Formulation}). We employ a plug-and-play strategy integrated with graph-enhanced reasoning prompt to execute DocRE (Section \ref{subsec: Decomposed Extraction}, \ref{subsec: Verifier}, \ref{subsec: Graph-enhanced Plug-and-Play}) under the few-shot scenario.

\subsection{Task Formulation}
\label{subsec: Task Formulation}
Given an input document $x$ with $l$ tokens $x_1, x_2, . . . , x_l$, a relation extraction task is to predict structured
target $y$ from $x$ for the query entity pairs, $\mathcal{Q}=\{(e_1, e_2)_i\}^M_{i=1}$. $M$ is the number of query pairs in the document. $e_1, e_2$ are two pre-provided entities spanning from $x$.
The prediction target $y$ of RE consists of a set of triplets $(e_1, r, e_2)$, where $r \in \mathcal{R}$ is the semantic relation (e.g., "head of government") between the two entities.
Here $\mathcal{R}$ denotes a pre-defined relation type set.
%In the few-shot setting, we are given a small set of annotated samples $\{(x_i, y_i)\}^n_{i=1}$ that consists of $k$ samples per class to compose a $k$-shot setting.
In order to leverage generative LLMs for DocRE tasks, we reformulate DocRE tasks as a generation task, which models the probability for generating a consequential string $y$ of a collection of relation triplets, conditioned over a context string $\mathcal{C}$. The context string consists of a small number of $k$ linearized samples $\{(x_i, y_i)\}^n_{i=1}$, where $k << N$ in a few-shot setting, and $N$ is the total number of samples.

We wrap the input sample $X$ and ground truth $Y$ into a generative style prompt conditioned under the relation triplet reasoning explanation $E$. 
The explanation generation aims to bridge the gap between the pre-defined label type space and real-world descriptions of label types, enabling LLMs to leverage their capabilities in a more familiar context.

%, alleviating LLMs' confusion on label types. 
As shown in Figure \ref{fig: Framework}, a LLM takes in a sample document context and a list of entity mentions to generate the relation triplets including all possible mentioned entity pairs with an explanation per extracted triplet. 
%Given a lengthy context, a graph can be distilled by all extracted triplets after processing the context document.
Given a lengthy context, all extracted triplets could form a graph %for better understanding the context document
which further enhances missing relation triplet generation for LLMs. Since the triplets and explanations generated from LLMs are not fully trustworthy, we further propose to employ a verifier to examine whether any entity pair that should be associated with a certain relation type has been missing, discussed in Sec. \ref{subsec: Verifier}. 
%While the triplets and explanations generated from LLMs may not be entirely trustworthy, we suggest incorporating a verifier to assess whether the results have overlooked any entity pairs that should be linked to a specific relation type.

\subsection{Decomposed Extraction}
\label{subsec: Decomposed Extraction}
Despite using carefully designed prompts \cite{wadhwa2023revisiting,liu2023pre}, performing DocRE through LLMs without any samples presents a significant challenge. In a standard few-shot setting, it becomes essential to furnish LLMs with a small number of labeled samples to enhance results. 
%%%
The direct application of the prompting method from classification tasks encounters two primary challenges: (1) The complexity of generating answers from LLMs is heightened by a large label space. (2) In relation extraction, the involvement of entity mention types in a relation is influenced by constraints imposed by the relation type; for instance, the relation "year of birth" cannot occur between two entities of type "person."

Since ensemble prompts aims at extracting all possible triplets from all types of relations per document (applied by \cite{wadhwa2023revisiting}), LLMs face difficulties in both memorizing the relations in the list and effectively distinguishing between them.
In the following example, we provide the ensemble prompt including a prompt instruction, a given document context, and a collection of relation triplets.

\noindent\rule{\linewidth}{0.1mm}
{
\noindent {\color{Gray} \# Ensemble Prompt}

{
{\color{MidnightBlue} 
\noindent List the relation triplets among the entities marked in “**entity**” from the given context and provide the triplet semantic explanation. The relation labels are provided in the list:}

\noindent{\color{BrickRed} [‘head of government’, ‘basin country’, ‘country’,  ‘father’, … ]}

{\color{MidnightBlue} 
\noindent[Context]: 

\noindent[Relation Triplet]:
}

{\color{BrickRed}
\noindent(\textit{e1},head of government,\textit{e2}), [explanation] \textit{e1} is the governor of \textit{e2}.

\noindent(\textit{e3},country,\textit{e4}), [explanation] \textit{e3} is in country \textit{e4}.}}

\noindent\rule{\linewidth}{0.1mm}
The ensemble relation type list is provided in the prompt instruction. 
Experimentally, the generation prefers relations seen in the few-shot prompt samples. Factors such as the prolonged generation time required to enumerate all conceivable entity pairs across all relation types, coupled with constraints on the length of generated tokens hinder the diversity of generation. This limitation applies to both different entity pairs and various relation types. 
The aforementioned challenges contribute to a high missing rate, reflected in a low recall score of relation types, during the relation extraction process.

Two noteworthy points arise here: (1) LLMs lack label distinction, and (2) the inclusion of an ensemble label list in the prompt presents challenges in terms of LLMs' instruction comprehension.  
To tackle these issues, we break down the prompt into sub-prompts for each relation type, the so-called decomposed prompt, as shown in the example below. 

\noindent\rule{\linewidth}{0.1mm}
{
\noindent {\color{Gray} \# Decomposed Prompt}

{\color{MidnightBlue} 
\noindent Based on the context, assign the relation {\color{BrickRed}“basin country”} for possible entity pairs and entities are marked in "**entity**":

{\color{MidnightBlue} 
\noindent[Context]: 

\noindent[Relation Triplet]:
}

{\color{BrickRed}
\noindent 
(**Vltava**, basin country, **Czech Republic**) | Because the relation “basin country” means the object is a country that has drainage to/from or borders the lake subject or the river subject. The Vltava is the longest river in the country, Czech Republic.
}}}

\noindent\rule{\linewidth}{0.1mm}
Different from a list of relation types in the ensemble prompt, the decomposed prompt focuses on only one type each time, denoted in the instruction part. As a result, the decomposition not only reduces the complexity of instruction by shortening its length but also saves prompt space to load more few-shot samples.
\begin{figure*}[ht!]
    \centering
    \includegraphics[width= \textwidth]{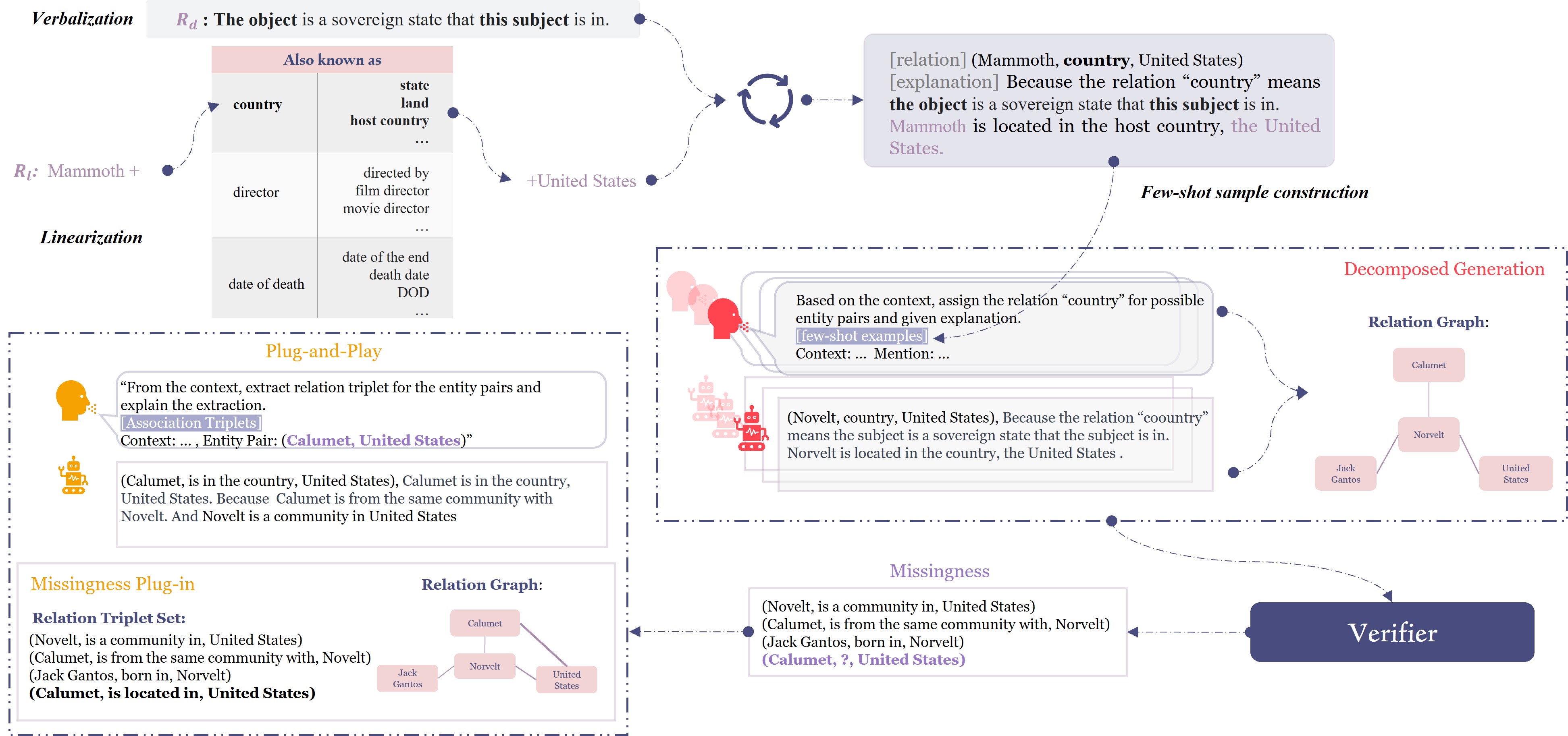}
    \vspace{-0.1in}
    \caption{Decomposed Plug and Ensemble Play Framework.}
    \label{fig: Framework Details}
\end{figure*}

\begin{figure*}[ht!]
    \centering
    \includegraphics[width= \textwidth]{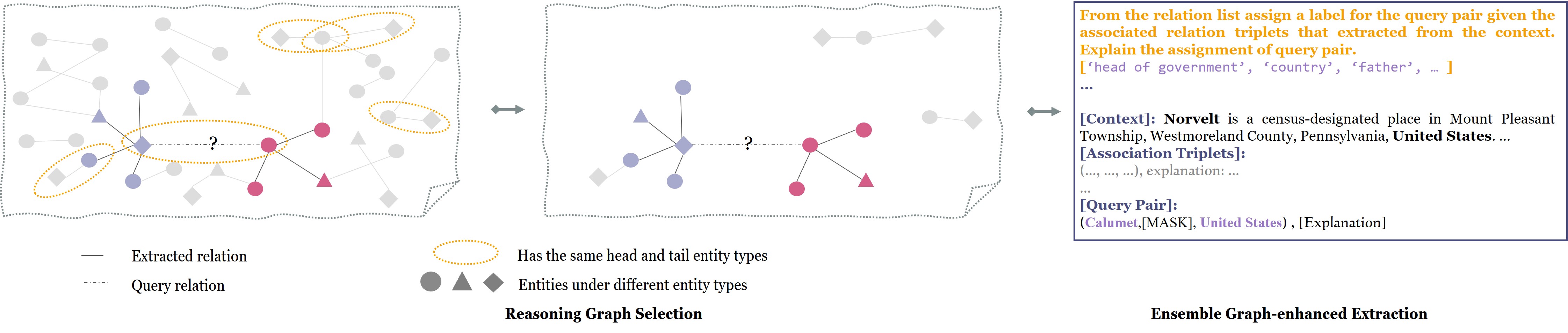}
    \vspace{-0.1in}
    \caption{Reasoning graph selection and Ensemble prompt for plugging in missing pairs}
    \label{fig: ensemble prompt}
\end{figure*}
\noindent\textbf{Type-injection Explanation} In-context learning \cite{dong2022survey} enhances the capability of LLMs by adding a few annotated examples written in natural language. However, it exhibits sensitivity to the template, order, and selection of these examples. 
 Given the limitation where a relation type's sparse words may not sufficiently convey how to associate with entities in a generated pair, 
%Considering limited words in a relation type cannot tell well about how to associate with the entities in a generating pair, 
we make use of Chain-of-Thought reasoning \cite{wei2022chain} with natural-language-described explanation for generating the triplet. We propose to inject the entity and relation type information into the explanation to enhance the type identification capability of LLMs especially when they are facing multi-grained relation types like "country" and "basin country". Type injection increases the verbosity of triplet explanation for better interpretation of both relation type description and entity type. This involves reiterating the reasoning behind the generation and emphasizing type constraints. We leverage the WikiPage (e.g. ``Basin Country" \footnote{https://www.wikidata.org/wiki/Property:P205}) of the relation type to construct the explanation, where we employ two entries, \textit{Description} and \textit{Also known as}, to compose \textbf{verbalization} of type definition ($R_d$) with placeholders of both subject and object entities and \textbf{linearization} on specific entity words ($R_l$) with alternatives in \textit{Also known as}, as shown in Figure \ref{fig: Framework Details}.

On one hand, \textbf{verbalization} functions as the verbosity of relation type to enhance type understanding. We insert ``the subject" and ``the object" into the originally retrieved \textit{Description} (e.g. ``sovereign state that this item is in"), and verbalize it in natural language (e.g. $R_d$:``The object is a sovereign state that this subject is in."). 
On the other hand, \textbf{linearization} focuses on the coherence of both entity and relation types when rephrasing a triplet into a natural language interpretation. 
This requires the replacement of ``the subject" and ``the object" with specific $e1$ and $e2$ remains meaningful in the context. For example, given a triplet ``(Virginia, head of government, McAuliffe)", the relation type ``head of government" %(means that the object is the head of the executive power of the subject town, city, municipality, state, country, or other governmental body) 
only applies between a person and a town/state/country. In this sense, $e1$ should be a person while $e2$ should be a town/state/country. 
Furthermore, we exploit the \textit{Also known as} entry to modify the word usage in \textit{Description} which increases sentence diversity accessible to LLMs. 
As shown in the example in Figure \ref{fig: Framework Details}, (Mammoth, country, United States) is interpreted as ``Mammoth is located in the host country, the United States."

\subsection{Verifier for Plugging} 
\label{subsec: Verifier}
The decomposed prompt not only increases type generation variance but also saves model computation time by reducing instruction space and extraction complexity. 
Consequently, this allows for an increase in new tokens and generation diversity within the constraints of the limited processing length of LLMs.
We further conduct a self-verification process on LLMs' generation. Similar to humans, the language models as discussed in \cite{pan2023automatically} can achieve refinement with persuading confidence under self-investigation and self-correctness.  

We first clean the generation by shrinking the repetition, incompleteness, and irrelevant entity. Repetition in triplet generation can be alleviated by hyper-parameter tuning but still exists. Incompleteness could be caused by the limited number of newly generation tokens of LLMs or misunderstanding of the template in the few hot samples. Neither an incomplete triplet nor an incomplete explanation should be discarded. Besides, we also face the irrelevant entity issue where a triplet contains a subject or object out of the entity mention list. After pruning the decomposed generation, we further filter out the outliers under each type. Specifically, incompleteness cases are observed by outlier detection. Moreover, outliers also include incoherent verbalization or linearization.
%Resolved: The following sentence is vague / the word "under" is used twice in a single sentence here which adds to the confusion. 
In constructing the few-shot sample template, we operate on the assumption that generating under a single type should align, forming a cluster of this type. The outliers are those far from the centroid of the type cluster with fewer neighbors.
We encode the explanation by a trivial pre-trained language model, e.g., BERT \cite{devlin2018bert}. 
% Resolved: Repeated use of the word "lowest" in the next sentence also leads to confusion and vagueness. Are they different lowest? Same lowest? Why the repetition. Do you really need the parenthetical lowest?
The outliers can be detected as the ones (Lowest K) that have the lowest neighbor density: $I_{outlier}=\text{LOF}(\text{Encoder}(exp_i)), exp_i \in E$, where $\text{LOF}$ is the function of estimating local outlier factor \cite{breunig2000lof} and implemented in Scikit-learn\footnote{https://scikit-learn.org/stable/index.html}.

%Resolved: Read and ensure correctness of the LAST sentence in the following paragraphs. is it "existing in Verifier" or "existing in the document"?
Another design in \textit{Verifier} is to identify the entity pair missed in the generation.  Given the entity mention pair list in the dataset, $Verifier_i= \{(e1_i, e2_i)\}$, if there is a $(e1_i, \_, e2_i) \in y_i$. \textit{Verifier} stores the set of entity pairs of a document that should be associated with one or more relation types. A missing entity pair, $(e1_{miss}, e2_{miss})$ refers to a pair not appearing in the generation associated with a relation but existing in query pair list $\mathcal{Q}$. 

\subsection{Graph-enhanced Plug-and-Play} 
\label{subsec: Graph-enhanced Plug-and-Play}
Missingness detection is crucial to the proposed plug-and-play regime, including decomposed individual relation extraction, missingness verification for each document, plugging the missing pairs into a few-shot DocRE prompt, and replaying the extraction to fill in their relation slots. This section presents the plugging prompt of relation extraction for missing pairs, under association triplets, the so-called reasoning relation graph. In terms of association, we assume that extraction on a query pair (a missing pair) should mostly benefit from the triples that share an entity with the query pair or that have the same subject and object types.

%Resolved: Please review my modification of the first sentence below. Is my change correct. I changed "whole generating tgriplets" to "all generated triplets".
\noindent\textbf{Reasoning Graph Selection.} We first distill the decomposed generation to save prompt-token cost and computational time by selecting the association sub-graph rather than employing all generated triplets.
% Resolved: Please check my minor modification of term in the next sentence
The association sub-graph, $G^a$,  comprises the triplets with entities that are shared with the missing pair and those whose entity types are common with the missing pair. 
We denote $G^a=\{(e1_i,r_i, e2_i)\}$, where $e1_i = e1_{miss}$ or $e2_i = e2_{miss}$ or $\{\text{type}(e1_i), \text{type}(e2_i)\} = \{\text{type}(e1_{miss}), \text{type}(e2_{miss})\}$. $\text{type}(\cdot)$ means entity type.

\noindent\textbf{Ensemble Graph-enhanced Extraction.} After obtaining $G^a$, we augment it as the prompt for reasoning the relation label of the missing pair. as shown in Figure \ref{fig: ensemble prompt}. The ensemble prompt contains all relation labels for the extraction of missing pairs and the association triplets, $G^a$, regarded as the reasoning thoughts so-called graph-of-thoughts in this paper. 
%Resolved: Is it "associated graph reasoning" or "reasoning on associated graph"? I changed to the latter .  Please review.
We ask the model to fill in the \textit{MASK} and \textit{Explanation} based on the understanding of context and reasoning on the association graph.

\section{Experiment}
\subsection{Setup}
\noindent\textbf{Dataset}
One of the most widely used benchmarks in this area is the DocRE \cite{yao2019docred}, which adopts a recommend-revise annotation scheme so as to have a large-scale annotated dataset. We evaluate our Graph-DPEP method on RE-DocRE dataset\cite{tan2022revisiting}, which originated from the DocRED dataset but contains a revision of 4,053 documents that resolved various problems including incompleteness, logical inconsistencies, and coreferential errors. 

We have used in total 97 relation types with 96 relations from RE-DocRE and an added 'NA' type.  
The total number of samples is 4,053 (3,000/500/500 for the train/dev/test split). In average, RE-DocRE contains 198.4 words/doc, 19.4 entities/doc, 28.1/34.6/34.9 triplets/doc of train/dev/test splits. 
%RE-DocRE contains 96 relations. Plus 'NA' type we have the generation on few-shot DocRE with 97 relation types. 

\noindent\textbf{ICL Data Selection.} In-context learning is known to be sensitive to the choice and even ordering of the demonstration examples \cite{perez2021true,lu2021fantastically}. Inspired by VOTE-K sample selection in \citet{su2022selective}, we construct a graph by applying k-NN over the sentence-transformer-embedded document contexts of the training set to select diverse and representative prototype samples, $\mathcal{S}$ (e.g. 1500). 
% \section{Sample Selection}
% \label{sec: sample selection}
Considering the two-stage prompting design in Graph-DPEP, we select $n$ samples with the occurrence of a specific relation type during the decomposed generation, namely $n\cdot|\mathcal{R}|$ in total. We randomly select $n$ samples in the pool, $\mathcal{S}$, for the ensemble graph-enhanced extraction.

\noindent\textbf{Few-Shot Setting.} We evaluate the impact of different number of query entity pairs in a document. Given different density of query pairs, we categorize the document into sparse ($\#\text{query pairs} \leq
 20$), normal($ 20 < \#\text{query pairs} \leq 40$), and dense ($40 \leq \# \text{query pairs}$) groups. We sample $k$ training samples for each relation type to construct a $k$-shot training set. The value of $k$ varies but should be within a maximum length limitation. We randomly select 50 samples of each group to conduct various $k$-shot settings (2/3/5-shot).

\noindent\textbf{Evaluation Metrics.} We adopt metrics designed in \citet{jiang2024genres}: (1) Micro-/Macro-F1 scores, assessing the task performance. (2) Topical Similarity (TS) score, measuring the information abundance of the extracted triples compared to the context. (3) Uniqueness Score (US), assessing the diversity of extraction and highlighting the importance of extracting varied and distinct relation types. (4) Factualness Score (FS), testing the alignment with information in the context to address hallucinations. (5) Granularity Score (GS), testing whether the extracted triplet can be further split into sub-triplets with more fine-grained types. (6) Completeness Score (CS), evaluating how comprehensively the extracted triples cover the information present in the context.

\noindent\textbf{Baseline.} Generative few-shot DocRE with LLMs interaction is still in an early stage of research. Most recent works address sentence-level DocRE along with the advent of in-context learning. 
%We conduct rigorous comparison experiments to illustrate the benefits of Graph-DPEP. 
COT \cite{tan2022revisiting} is the first study on adapting chain-of-thoughts reasoning on sentence-level relation extractions. We replicate the prompt instruction interpreted in this work, which is regarded as the ensemble prompt since all types are employed. To make a fair comparison, we adopt Llama2 as the backbone LLMs rather than GPT3 \cite{brown2020language} reported in the paper. CODEIE \cite{li2023codeie} is another ensemble prompt, which requires the generation in the python-code style, replicated by an open-source LLM, Code Llama. We obtain ChatGPT from OpenAI with the model API, {\fontfamily{qcr}\selectfont gpt-3.5-turbo-instruct}.

\begin{table*}[]
\small
\centering
\renewcommand\arraystretch{1.3}
\resizebox{1.0\linewidth}{!}{% <-
\begin{tabular}{cccccccccc|cccccc}
\toprule
                      & \multicolumn{3}{c}{\textbf{Sparse}} & \multicolumn{3}{c}{\textbf{Normal}} & \multicolumn{3}{c}{\textbf{Dense}} & \multicolumn{6}{c}{\textbf{All}}    \\ \hline
Metrics                      & \multicolumn{9}{c}{Micro-F1} 
& \multicolumn{1}{c}{Micro-F1}
& \multicolumn{1}{c}{TS}
& \multicolumn{1}{c}{US}
& \multicolumn{1}{c}{FS}
& \multicolumn{1}{c}{GS}
& \multicolumn{1}{c}{CS}\\ 
\multicolumn{1}{c}{\#shot}               & \multicolumn{1}{c}{2-shot}  & \multicolumn{1}{c}{3-shot}  & \multicolumn{1}{c}{5-shot} & \multicolumn{1}{c}{2-shot}  & \multicolumn{1}{c}{3-shot}  & \multicolumn{1}{c}{5-shot}& \multicolumn{1}{c}{2-shot}  & \multicolumn{1}{c}{3-shot}  & \multicolumn{1}{c}{5-shot}  & \multicolumn{6}{c}{3-shot} \\ \hline
\multicolumn{11}{l}{Ensemble Prompt}            \\ \hline                              
COT \cite{tan2022revisiting}                 & 8.22    & 11.83   & 13.41   & 8.33    & 15.21   & 16.66 & 6.66    & 7.56   & 8.21   & 7.39  & 7.03 & 80.23 & 29.57 & 30.07 & 8.98 \\
CODEIE \cite{li2023codeie}              & 4.34    & 6.99    & 7.23    & 4.04    & 6.32    & 7.77  & 6.49    & 6.88   & 7.84    & 7.23   & 7.24 & 82.30 & 24.66 & 26.11 & 11.04 \\ 
ChatGPT     & 21.93  & 34.22  &   41.08
& 32.29 & 36.25 & 37.33 
& 19.11 & 20.38 & 22.06
& 21.19   & 15.92 & \textbf{86.98} & \textbf{94.75} & 49.27 & 35.38  \\ \hline
\multicolumn{11}{l}{Decomposed Prompt}          \\ \hline        
Graph-DPEP on Llama2   & 25.88   & 37.03   & 41.77   & 36.09   & 38.81   & 42.00   & 22.93   & 24.56  & 27.01  & 28.95  & 14.39 & 85.77 & 34.47 & 59.99 & 37.32 \\
Graph-DPEP on Mistral  & 27.27   & 37.64   & 43.31   & 33.55   & 39.27   & 42.93   & 24.08   & 26.15  & 28.29  & 30.18  & 15.25 & 83.48 & 36.81 & 61.20 & 38.28 \\ 
Graph-DPEP on Llama3   & 27.58   & 37.99   & 44.74   & 36.09   & 41.01   & 43.43   & 24.03   & 27.56  & 29.94  & 32.07  & 15.66 & 86.91 & 43.98 & \textbf{61.78} & 42.09 \\ 
\hline
\multicolumn{11}{l}{Decomposed Prompt + Graph-enhanced Plug-and-Play}    \\ \hline                                            
Graph-DPEP on Llama2   & 26.27   & 39.66   & 42.06   & 35.04   & 37.89   & 43.47  & 23.71  & 24.78  & 28.10  & 29.27  & \textbf{17.01} & 82.93 & 35.20 & 57.17 & 38.28 \\
Graph-DPEP on Mistral  & 29.93   & 40.37   & 42.58   & 36.02   & 41.24   & 44.85   & 26.45   & 27.31  & 29.05   & 33.01  & 15.02 & 83.85 & 38.29 & 57.45 & 39.39 \\ 
Graph-DPEP on Llama3  & \textbf{30.11}   & \textbf{42.88}   & \textbf{43.27}   & \textbf{37.20}   & \textbf{41.63}   & \textbf{44.89}   & \textbf{28.02}   & \textbf{30.25}  & \textbf{31.77}   & \textbf{35.55}  & 16.42 & 86.47 & 46.88 & 60.05 & \textbf{43.75} \\\hline
\toprule
\end{tabular}
}
\caption{\label{tab: Main Results}
Performance of baselines and Graph-DPEP with different variants of prompt design. A higher TS indicates close alignment with the document's topical content. A higher US indicates greater diversity. A high FS signifies the portion of triples that are factually consistent with the document text. A lower GS indicates that the triples are too coarse-grained to be broken down further. A higher CS indicates the range of the document information captured by the extraction.
}
\end{table*}
%\noindent\textbf{Evaluation}

\subsection{Results}

% Resolved: I removed "only" because it is implied by the context here. If you want to add it, you need to use phrases like "Ensemble-only" and "Decomposed-only" which is still awkward but correct as an adjectives for "Prompt"...

\subsubsection{Ensemble Prompt vs. Decomposed Prompt} As shown in Table \ref{tab: Main Results}, the comparison made in the upper two groups presents the superiority of decomposed prompt when handling intensive type space, specifically from the performances of COT and Graph-DPEP on Llama2. The worst performer with the lowest Micro-F1 scores, CODEIE, reveals the drawbacks of code-style generation when dealing with lengthy context and intensive extraction. ChatGPT can obtain decent results under the ensemble group but is still slightly worse than Graph-DPEP. 
%For clarity, I made a correction in this last sentence.
Even in the most challenging scenario, i.e., the dense group, Graph-DPEP with 5-shot can obtain an 14.36\% gain when comparing ChatGPT with Graph-DPEP on Llama3. The task's best performers always come from the bottom group, which indicates the success of the decomposed-plug and ensemble-play framework.

Apart from the Micro-F1 score in Table \ref{tab: Main Results}, TS, US, FS, GS, and CS also indicate the observations that: (1) the decomposed-plug and ensemble-play framework shows their capability to interpret the topics of the text; (2) Decomposed prompt truly increase the diversity of relation extraction since scores in this group almost outperform the others, while the ensemble stage tends to generate triplets with frequent relations which causes the hurt on diversity; (3) the hallucination is alleviated with the complete two-stage framework. (Since the fact checker in FS is implemented by ChatGPT, the FS of ChatGPT is the best score.); (4) high GS and CS on Graph-DPEP indicate its capability of capturing more complete information in the document context.

In Table \ref{tab: macro evaluation}, we evaluate the average performance under each relation type by macro-precision, macro-recall, and macro-F1 scores on the entire dataset. The precision score in the decomposition group is far beyond that of COT and CODEIE, which validates the importance of not only type decomposition but also type injection with verbalization and linearization. 
% Resolved: Following what you said, I changed it to graph-of-thoughts.
Graph-of-Thoughts reasoning on missing pairs after decomposition generation further strengthens Graph-DPEP's performance leading it to surpass all baselines, including the powerful rival, ChatGPT. 
Compared with  ChatGPT, COT, and CODEIE, the outstanding macro-recall scores of Graph-DPEPs show their better comprehensive understanding of the context document with richer extracted triplets. 
In summary, the entire Graph-DPEP can overcome the predicted bias that appeared in the ensemble group of models witnessed by the impressive increase of the precision scores and compensate for the buried pairs excluded in the decomposition generation to bring recall gains, almost 3.10\% by Graph-DPEP on Llama3 compared with its variants, Graph-DPEP on Llama3 \textit{with only decomposed prompts}.

\begin{table}[]
\small
\centering
\renewcommand\arraystretch{1.3}
\resizebox{1.0\linewidth}{!}{% <-
\begin{tabular}{@{}llll@{}}
\toprule
 \textbf{Models}                                            & \textbf{Macro-Rec} & \textbf{Macro-Pre} & \textbf{Macro-F1} \\ \midrule
ChatGPT                                      & 37.31              & 20.31              & 26.30             \\
COT \cite{li2023codeie}                                           & 16.02              & 17.85              & 16.88             \\
CODEIE \cite{li2023codeie}                                       & 8.30               & 15.45              & 10.79             \\
Graph-DPEP on Llama2 \textit{w. only dec. prompt}  & 58.76              & 21.34              & 31.31             \\
Graph-DPEP on Mistral \textit{w. only dec. prompt} & 61.17              & 25.30              & 35.79             \\
Graph-DPEP on Llama3 \textit{w. only dec. prompt} & 62.17              & 30.17              & 40.62             \\

Graph-DPEP on Llama2                         & 62.49              & 22.82              & 33.43             \\
Graph-DPEP on Mistral                        & 64.69              & 27.99              & 39.07             \\
Graph-DPEP on Llama3                        & \textbf{65.27}              & \textbf{31.46}              & \textbf{42.45}             \\\bottomrule
\end{tabular}}
\caption{\label{tab: macro evaluation}
Macro precision, recall, F1 scores for all baselines and Graph-DPEP variants.
}
\end{table}
\begin{figure}[ht!]
    \centering
    \includegraphics[width= 0.48\textwidth]{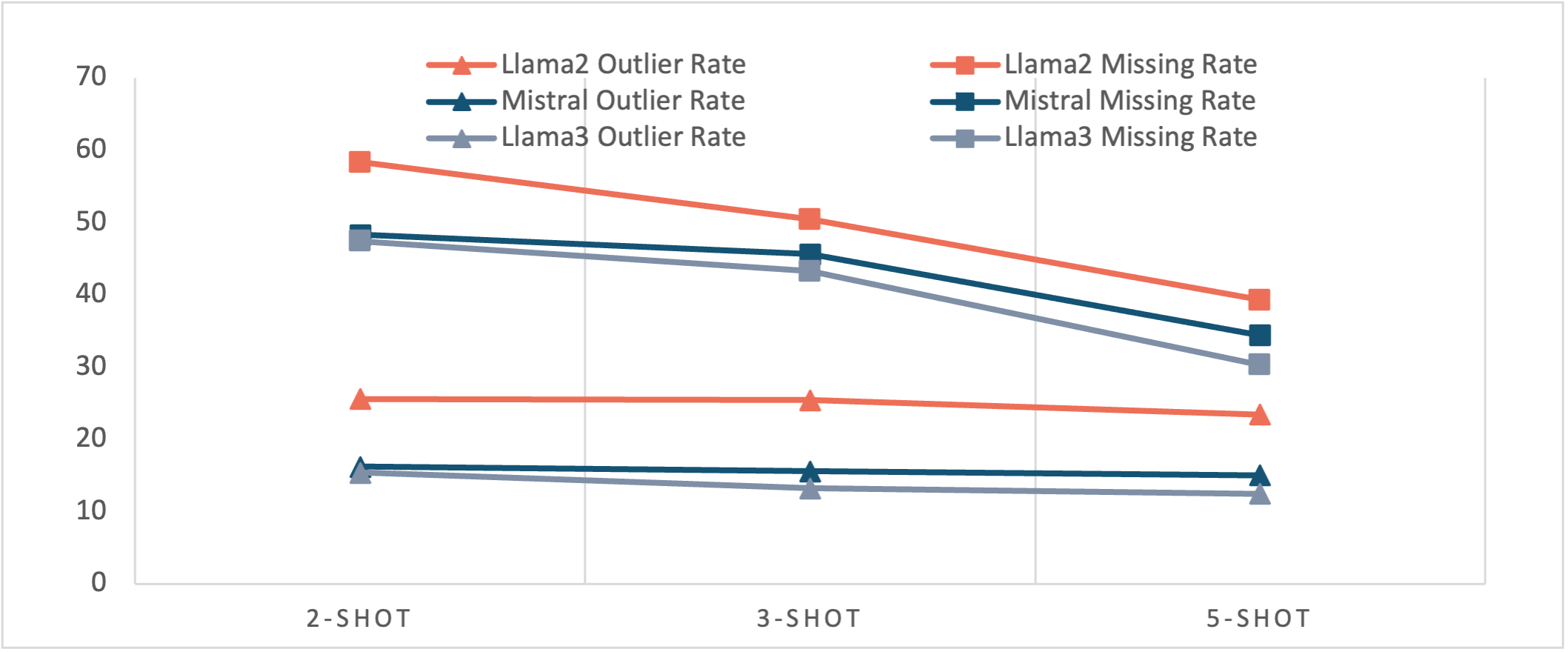}
    \caption{Outlier rate and missing rate of k-shot Graph-DPEP with Llama2 and Mistral.}
    \label{fig: outlier and minssingness}
\end{figure}

\subsubsection{Different LLMs' Impacts} %on Verifier
In Figure \ref{fig: outlier and minssingness}, we illustrate the results of the outlier rates and missing rates of Graph-DPEP with different LLMs. The outlier rate is given by $\frac{\#outlier}{\#generation\:triplets}$, while the missing rate is given by $\frac{\#missing\:pairs}{\#ground-truth\:pairs}$. The results show that Llama2 suffers more severely on both outliers (incompleteness and irrelevance) and missing pairs among all $k$-shot settings. 
%The Mistral outperforms Llama2 in every aspects. 
Another significant observation is that while the outlier rate rarely changes along with increasing shots from 2-shot to 5-shot, the missing rate can be alleviated by adding more few-shot samples. LLMs' abilities like instruction comprehension, volume and types of pre-training corpus, and generation-related hyperparameter-tuning account for the outbreak of outliers.
%
%On the contrary, appending more few-shot samples does alleviate the missing pair issue, witnessed through the decrease from 2-shot to 5-shot in both Llama2 and Mistrial settings.

%Resolved: Changed to "Graph-of-Thoughts". Please review for consistency across the board to the extent possible.
\subsubsection{Graph-of-Thoughts Reasoning}
Rather than employ the ensemble prompt in COT \cite{tan2022revisiting} for the missing pairs, we identify the most relevant associations with the query pair in the generated graph-of-thoughts. This approach casts DocRE problem into a graph-of-thoughts reasoning scheme. Figure \ref{fig: graph} illustrates the Graph-enhanced plug-and-play process with different graph-of-thoughts reasoning schemes, with either the overall graph or the selected sub-graph. The results show the average scores, and their corresponding variances for each $k$-shot setting of three groups, i.e., Sparse, Normal, and Dense.

Experiments are conducted by randomly selecting few-shot samples 3 times from the annotation pool (M=1500 samples) to obtain the variance scores. Considering the generation token limitation, the annotation pool of the overall graph category only contains sparse samples, otherwise, we cannot load 3-shot or 5-shot samples into the prompt. That is one of the reasons why the Sparse group exploiting the overall graph achieves slightly higher than the selected association sub-graph. Another reason is that the association graph in the Sparse group is still sparse, and lacks a coherent and informative graph for reasoning purposes. 

From the last two columns in Figure~\ref{fig: graph}, scores in Selection with sub-graph-of-thoughts tend to overtake the Overall graph-of-thoughts as expected. Thanks to the lighter graph, the association sub-graph pool can enjoy greater diversity by loading samples from not only Sparse but also Normal and Dense which benefits from the generalizability and robustness of LLMs under few-shot scenarios. Therefore, the sub-graph-of-thoughts can surpass the overall graph-of-thoughts under the 3-shot and 5-shot settings.

\begin{figure}[ht!]
    \centering
    \includegraphics[width= 0.48\textwidth]{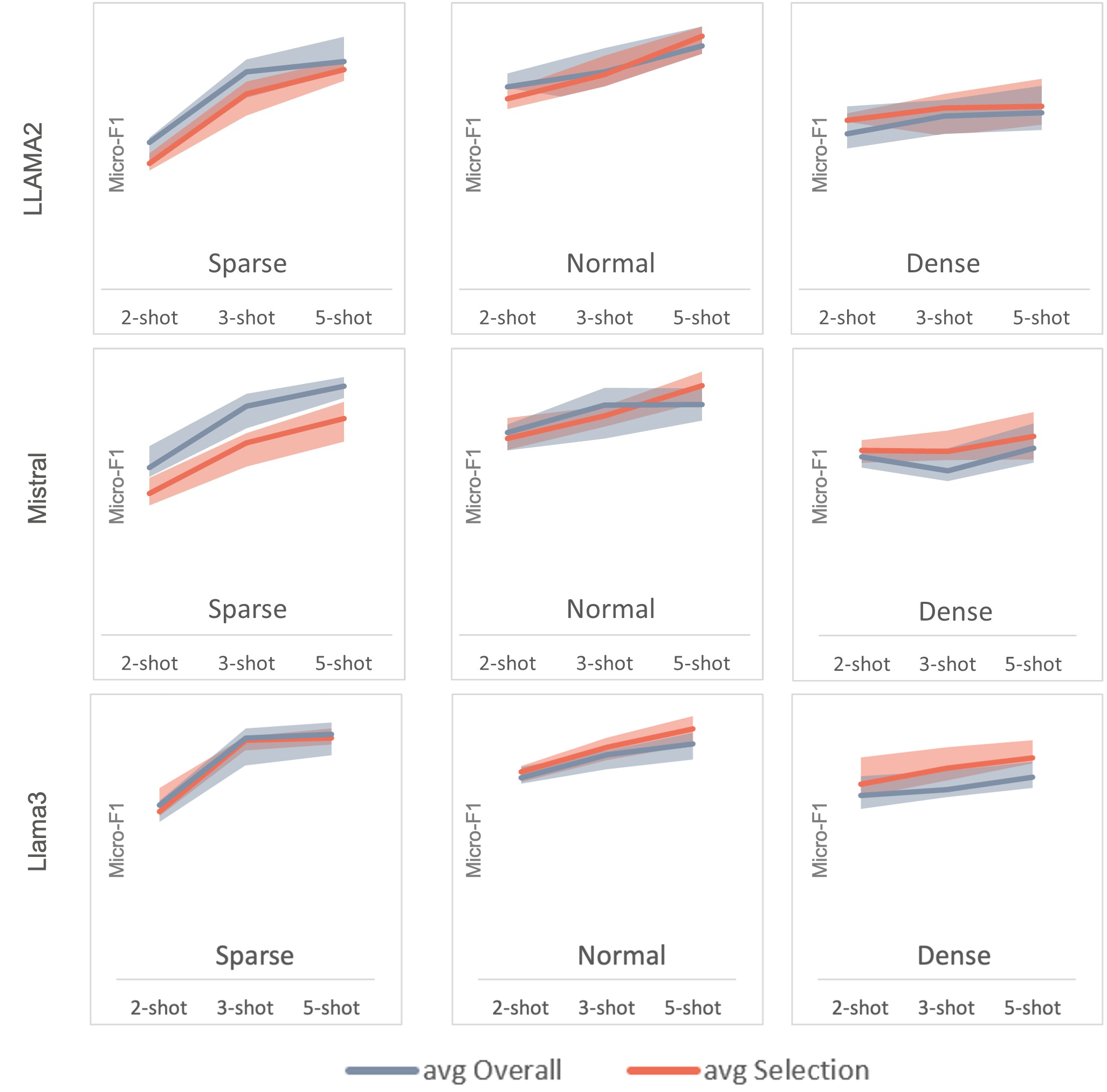}
    \caption{Overall Graph vs. Selection Graph}
    \label{fig: graph}
\end{figure}

\section{Case Study}
\label{sec: case study}
Given the challenge of transitioning traditional extraction tasks into text-to-text generation, we offer a detailed case study in Figure \ref{fig: example} and performance comparison in Table \ref{tab: case study} to highlight Graph-DPEP's value in DocRE. We evaluate the complexity of extracting complete relations from lengthy contexts by randomly selecting 50 samples and categorizing their query pairs into 4 reasoning types (percentage in 1st column, definition is after name entry), assessing them across three models: SAIS \cite{xiao2021sais}, COT\cite{tan2022revisiting}, and Graph-DPEP.

\textbf{(1) Pattern Recognition}. SAIS, with its multi-task architecture, delivers entity type, relation extraction, and evidence sentence indexes. Graph-DPEP, despite only achieving extraction at the decomposed stage, maintains competitive F1 score at 68.5\% compared to SAIS, while COT scores 20.1\% and Graph-DPEP achieves 35.4\%. This highlights how Graph-DPEP, by leveraging decomposed-plug and ensemble-play prompting LLMs, outperforms general COT reasoning methods in DocRE.

\textbf{(2) Logical Reasoning}: The example underscores the importance of reasoning through the intermediary entity 'X-Files'. While COT fails to extract any relation, Graph-DPEP accurately utilizes association triplets reasoning to extract the relationship. The association graph highlights the intermediary entity's crucial role in linking relation labels between query entities. However, significant performance differences exist among models: 80.2\% in SAIS, 16.0\% in COT, and 37.1\% in Graph-DPEP.

\textbf{(3) Coreference Reasoning}: In the example, 'Dwight Tillery' and 'Tilley' refer to the same person. Graph-DPEP handles relations only at the decomposed stage, whereas COT misses due to failing in coreference resolution. Decomposed prompts reveal the model's ability to recognize entities with different surface forms in lengthy contexts. Performance varies significantly: 55.1\% in SAIS, 6.3\% in COT, and 24.1\% in Graph-DPEP. The gap between SAIS and Graph-DPEP narrows here.

\textbf{(4) Common-sense Reasoning}. This example highlights LLMs' noise sensitivity and the importance of self-calibration in common-sense reasoning. Despite potential noise in Graph-DPEP's triples, LLMs can accurately infer relations with the help of an association graph. Notable performance variations (49.7\% in SAIS, 6.6\% in COT, and 20.2\% in Graph-DPEP) underscore the need for advanced prompting designs to ensure efficient and accurate LLM inference.

Traditional PLM extraction models like SAIS struggle with effectiveness and efficiency in complex, large-scale data scenarios, while LLMs generate results without tuning for smaller tasks due to their exceptional natural language understanding and generalizability. Compared with CoT, Graph-DPEP is advancing LLM generation for DocRE, narrowing the performance gap with PLMs like SAIS. Compared with SAIS, Graph-DPEP attains a recall score with a small gap (67.4\% vs. 47.8\%) but lower precision score (27.5\% vs. 69.9\%) due to noise extraction, highlighting the importance of calibrating generation for LLMs in specific tasks.

Decomposed generation in Graph-DPEP extracts diverse relations for a query pair, providing a wider range than COT. The association graph deduces missing relations, including common ones like 'country', 'developer', 'platform', and 'located in the administrative territorial entity'. However, fine-grained relation types may overlap, like 'country' and 'located in the administrative territorial entity'. Despite Graph-DPEP's ability to extract both rare and frequent relations, noise may occur, as seen in the incorrect extraction of (Szczytno County, 'capital', Szczytno), revealing limitations in effectively filtering improper entity pairs in complex cases.

\begin{table}[]
\small
\centering
\renewcommand\arraystretch{1.3}
\resizebox{0.9\linewidth}{!}{% <-
\begin{tabular}{lllll}
\hline
                                        & \textbf{Models}     & \textbf{Pre}   & \textbf{Rec}  & \textbf{F1}   \\ \hline
\multirow{3}{*}{Pattern Recognition}    & SAIS       & 72.1  & 65.2 & 68.5 \\
                                        & COT        & 20.3  & 21.2 & 20.1 \\
                                        & Graph-DPEP & 27.39 & 49.9 & 35.4 \\ \hline
\multirow{3}{*}{Logical Reasoning}      & SAIS       & 81.2  & 79.3 & 80.2 \\
                                        & COT        & 14.3  & 18.2 & 16.0 \\
                                        & Graph-DPEP & 29.5  & 49.9 & 37.1 \\ \hline
\multirow{3}{*}{Coreference Reasoning}  & SAIS       & 51.3  & 59.3 & 55.1 \\
                                        & COT        & 10.2  & 4.5  & 6.3  \\
                                        & Graph-DPEP & 18.3  & 35.3 & 24.1 \\ \hline
\multirow{3}{*}{Common-sense Reasoning} & SAIS       & 49.2  & 50.2 & 49.7 \\
                                        & COT        & 8.7   & 5.3  & 6.6  \\
                                        & Graph-DPEP & 15.1  & 30.4 & 20.2 \\ \hline
\multirow{3}{*}{Overall}                & SAIS       & 69.9  & 67.4 & 68.6 \\
                                        & COT        & 7.3   & 10.3 & 8.6  \\
                                        & Graph-DPEP & 27.5  & 47.8 & 34.9 \\ \hline
\end{tabular}}
\caption{\label{tab: case study}
Performance comparison between traditional PLM extraction (SAIS), in-context learning generation (COT), and Graph-DPEP.
}
\end{table}

\begin{figure}[ht!] 
    \centering
    \includegraphics[width= 0.5\textwidth]{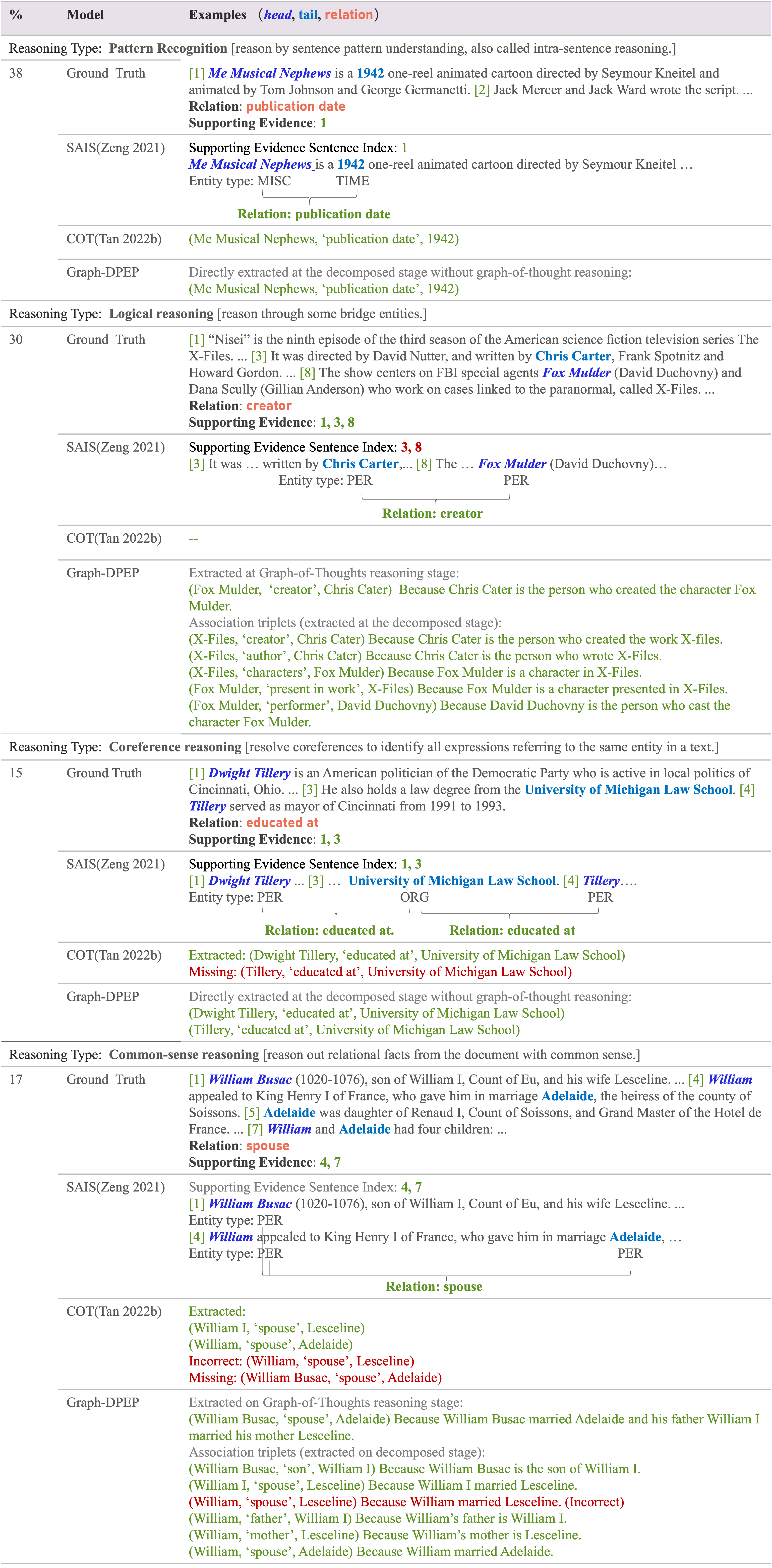}
    \caption{Case study for performance comparison between traditional PLM extraction (SAIS), in-context learning generation (COT), and Graph-DPEP.}
    \label{fig: example}
\end{figure}
\section{Conclusion}
In this work, we introduce a self-calibrating decomposed-plug and ensemble-play framework with graph-of-thoughts reasoning to address few-shot DocRE. 
%Resolved: The following sentence needs a lot of work. Please review.
Relation triplet generation along with an explanation(s) as the reasoning thoughts facilitate LLMs to generate proper natural language responses about a structural task output. 
Type decomposition and relation-entity type injection in explanation further assist LLMs in differentiating relation types when the type space is dense with complex granularity. 
The verifier’s calibration supports concise graph-of-thoughts in an ensemble prompt to save performance on query pairs missed in the decomposed generation stage. 
%Resolved: Review this change: "few-shot generation DocRE" --> "few-shot generative DocRE" 
Experiments on the benchmark exhibit the superiority of the proposed Graph-DPEP in few-shot generative DocRE.
\section{Discussion}
%\noindent\textbf{Annotation} During decomposed generation, we ask for human annotation on few-shot samples with verbalization and linearization. 
%
%To further ensure the quality of explanation annotation, we ask ChatGPT to correct the grammar issues if any, but not rephrase sentences because of consistency with retrieval alternatives in the entry, \textit{Also known as}. 

\noindent\textbf{Broader Impacts.}
The decomposed-plug and ensemble-play framework can be applied to any few-shot text generation task under dense label space with complex label granularity and diversity. 
Plus the graph-of-thoughts reasoning, The key advantages of the framework are two-folds:
(1) Challenge Divisibility. Thanks to the decomposition design, the framework is flexible to tackle generation on intensive label space tasks, alleviating the impact of long-tail distribution and retaining recall on infrequent labels. (2) Controllability. Pruning and missingness inspection in the verifier remedy different specific noise types, exerting greater control over the quality of graph-of-thoughts selection to complete the plug-and-play scheme.

\noindent\textbf{Limitations and Future Works.}
One major limitation of our framework comes from the type injection in the explanation of each relation type with verbalization and linearization, which costs human annotations and retrieval resources. Even though current annotation for a few-shot setting is manageable and cost-effective, the concern regarding expensive annotation will be amplified when scaling up the label size from hundreds to thousands. 
Therefore, it is necessary to conduct further study for upgrading the decomposition from one-by-one into group-by-group type decomposition. 
When considering the different granularity of relation types, similar fine-grained types should be distinct under a single coarse-grained type group.
After investigating the label structure, we may save annotation effort and decomposed generation time by group-level decomposition.

%%
%% The next two lines define the bibliography style to be used, and
%% the bibliography file.
\bibliographystyle{ACM-Reference-Format}
\bibliography{main}

%%
%% If your work has an appendix, this is the place to put it.
\appendix

\newpage
\section{LLMs}
\label{sec: LLMs and baselines}
\noindent\textbf{LLMs Compared.} We discuss the LLM models either used in Graph-DPEP or for replicating the baseline prompt: 

(1) Llama2~\cite{touvron2023llama} and Llama3 are two of the best performance open-source LLMs. We obtain the model, {\fontfamily{qcr}\selectfont Llama-2-7b-hf} via huggingface hub\footnote{https://huggingface.co/meta-llama/Llama-2-7b-hf} and {\fontfamily{qcr}\selectfont Meta-Llama-3-8B} via huggingface hub\footnote{https://huggingface.co/meta-llama/Meta-Llama-3-8B}. 

(2) Mistral \cite{jiang2023mistral} is one of the newest released LLMs outperforming Llama 2 on some benchmarks. It uses sliding window attention to exploit the stacked layers of a transformer to attend in the past beyond the window size, which saves half of the cache memory for inference without impacting model quality. We experiment with {\fontfamily{qcr}\selectfont Mistral-7B-Instruct-v0.2} \footnote{https://huggingface.co/mistralai/Mistral-7B-Instruct-v0.2} version Mistral.

(3) Code Llama, an open-access version of Llama 2 specialized in code tasks. We adopt it from {\fontfamily{qcr}\selectfont CodeLlama-7b-hf}\footnote{https://huggingface.co/codellama/CodeLlama-7b-hf} to replicate the baseline \cite{li2023codeie} prompt. 

All LLMs used in the experiments are configured to generate a maximum of at least 4096 tokens. That is one requisite for LLMs selection to deal with document-level extraction tasks.

\section{Decomposed Prompt Sample}
Based on the context, assign the relation “head of government” for possible entity pairs and entities are marked in "**entity**". To help you, I provide examples of relation “head of government”.  
Examples: 

[context]:**Alton** is a city on the **Mississippi River** in **Madison County** , **Illinois** , **United States** , about north of **St. Louis** , **Missouri** . The population was **27,865** at the **2010** census . It is a part of the **Metro - East** region of the **Greater St. Louis** metropolitan area . It is famous for its limestone bluffs along the river north of the city , for its role preceding and during the **American Civil War** , and as the home town of jazz musician **Miles Davis** and **Robert Wadlow** , the tallest known person in history . It was the site of the last **Abraham Lincoln** and **Stephen Douglas** debate in **October 1858** . The former state penitentiary in **Alton** was used during the **Civil War** to hold up to **12,000** **Confederate** prisoners of war .

[Relation]:
(**United States**, 'head of government', **Abraham Lincoln**) | Because relation “head of government” means the object is the head of the executive power of this suject, which can be A town, city, municipality, state, country, or other governmental body. Abraham Lincoln is a person. United States is a country. Abraham Lincoln is the presendent of the United States.

[context]:**Herzogenbusch** concentration camp ( , , ) was a **Nazi** concentration camp located in **Vught** near the city of ' **s - Hertogenbosch** , **Netherlands** . **Herzogenbusch** was , with **Natzweiler - Struthof** in occupied **France** , the only concentration camp run directly by the **SS** in western **Europe** outside **Germany** . The camp was first used in **1943** and held **31,000** prisoners . **749** prisoners died in the camp , and the others were transferred to other camps shortly before the camp was liberated by the **Allied Forces** in **1944** . After the war the camp was used as a prison for **Germans** and **Dutch** collaborators . Today there is a visitors ' center with exhibitions and a national monument remembering the camp and its victims . The camp is now a museum .

[Relation]:
Cannot find a pair.

\section{Graph-enhanced Ensemble Prompt Sample}
From the relation list assign a label for the query pair given the associated relation triplets that are extracted from the context. Explain the assignment of query pair.
[‘head of government’, ‘country’, ‘father’, … ]

[Context]: **Alton** is a city on the **Mississippi River** in **Madison County** , **Illinois** , **United States** , about north of **St. Louis** , **Missouri** . The population was **27,865** at the **2010** census . It is a part of the **Metro - East** region of the **Greater St. Louis** metropolitan area . It is famous for its limestone bluffs along the river north of the city , for its role preceding and during the **American Civil War** , and as the home town of jazz musician **Miles Davis** and **Robert Wadlow** , the tallest known person in history . It was the site of the last **Abraham Lincoln** and **Stephen Douglas** debate in **October 1858** . The former state penitentiary in **Alton** was used during the **Civil War** to hold up to **12,000** **Confederate** prisoners of war .

[Association Triplets]:
(**Miles Davis**, 'place of birth', **Alton**) | Because Alton is the home town of Miles Davis, which indicates Robert Wadlow was born in Alton.
(**Robert Wadlow**, 'residence', **Alton**)  | Because Alton is the home town of Robert Wadlow, which indicates Robert Wadlow is a residence in Alton.
(**Alton**, 'located in the administrative territorial entity', **Madison County**) | Because Alton is a city of Madison County.
(**Alton**, 'country', **United States**)  | Because Alton is a city of Madison County, Illinois, United States.
(**Robert Wadlow**, 'country of citizenship', **United States**)  | Because Alton is a city of  United States and Robert Wadlow is a residence in Alton, then Robert Wadlow has the citizenship of country, United States.
(**Miles Davis**, 'country of citizenship', **United States**)  | Because Alton is a city of  United States and Miles Davis's home toen is Alton, then Miles Davis has the citizenship of country, United States.
(**Alton**, 'located in the administrative territorial entity', **United States**) Because Alton is a city of United States.
(**Alton**, 'located in the administrative territorial entity', **Illinois**) Because Alton is a city of Illinois.

[Query Pair]: 
(**Robert Wadlow**, [MASK], **Alton**) | [Explanation]

[Answer]:
(**Robert Wadlow**, 'place of birth', **Alton**) | Because Alton is the home town of Robert Wadlow, which indicates Robert Wadlow was born in Alton.

\end{document}